\documentclass{article}

    \PassOptionsToPackage{numbers, compress}{natbib}

\usepackage[preprint]{neurips_2023}

\usepackage[utf8]{inputenc} 
\usepackage[T1]{fontenc}    
\usepackage{hyperref}       
\hypersetup{
	colorlinks   = true, 
	urlcolor     = blue, 
	linkcolor    = blue, 
	citecolor   = blue 
}
\usepackage{url}            
\usepackage{booktabs}       
\usepackage{amsfonts}       
\usepackage{nicefrac}       
\usepackage{microtype}      
\usepackage{xcolor}         

\usepackage{hhline}
\usepackage{amssymb}
\usepackage{array}
\usepackage{pifont}
\usepackage{enumerate}

\usepackage{xspace}
\usepackage{algorithm,algpseudocode}
\usepackage{algorithmicx}
\usepackage{stmaryrd}
\usepackage{bm}
\usepackage{multirow}
\usepackage{graphicx}
\usepackage{arydshln}
\usepackage{mathtools}
\usepackage{tabulary}
\usepackage{booktabs}
\usepackage{subfig}
\usepackage{comment}

\usepackage[textsize=tiny]{todonotes}

\usepackage{amsthm}
\newtheorem{theorem}{Theorem}

\newtheorem{lemma}{Lemma}

\ifodd 1

\newcommand{\com}[1]{\textbf{\color{blue} (COMMENT: #1)}}

\else

\newcommand{\com}[1]{}
\fi

\newcommand{\alg}{\texttt{DoE}\xspace}
\newcommand{\balg}{\texttt{DoE-bandit}\xspace}

\newcommand{\ci}{\textsc{CI}\xspace}
\newcommand{\ei}{G\xspace}
\newcommand{\CMAB}{{\normalfont \texttt{CMA2B}}\xspace}

\title{Cooperative Multi-agent Bandits: Distributed Algorithms with Optimal Individual Regret and Constant Communication Costs}

\author{
   Lin Yang\\
   Nanjing University\\
   \texttt{linyang@nju.edu.cn}\\
   \And
   Xuchuang Wang\\
    The Chinese University of Hong Kong\\
   \texttt{xuchuangw@gmail.com} \\
   \And
    Mohammad H.~Hajiesmaili\\
    University of Massachusetts Amherst\\
   \texttt{hajiesmaili@cs.umass.edu} \\
    \And
   Lijun Zhang\\
     Nanjing University\\
   \texttt{zhanglj@lamda.nju.edu.cn} \\
      \And
   John C.~S.~Lui \\
    The Chinese University of Hong Kong\\
   \texttt{cslui@cse.cuhk.edu.hk} \\
   \And
     Don Towsley\\
   University of Massachusetts Amherst\\
   \texttt{towsley@cs.umass.edu} \\
}

\begin{document}

\maketitle

\begin{abstract}
  Recently, there has been extensive study of cooperative multi-agent multi-armed bandits where a set of distributed agents cooperatively play the same multi-armed bandit game. The goal is to develop bandit algorithms with the optimal group and individual regrets and low communication between agents. The prior work tackled this problem using two paradigms: leader-follower and fully distributed algorithms. Prior algorithms in both paradigms achieve the optimal group regret. The leader-follower algorithms achieve constant communication costs but fail to achieve optimal individual regrets. The state-of-the-art fully distributed algorithms achieve optimal individual regrets but fail to achieve constant communication costs. This paper presents a simple yet effective communication policy and integrates it into a learning algorithm for cooperative bandits. Our algorithm achieves the best of both paradigms: optimal individual regret and constant communication costs.
\end{abstract}

\section{Introduction}

Recently there has been a surge of various online learning problems in distributed settings, where a set of agents perform individual learning algorithms to complete a common task and can cooperate with each other to improve the performance of the learning process. Distributed online learning is naturally motivated by a broad range of applications where computational resources are geographically distributed, and a group of machines has to communicate with each other to complete a common task cooperatively. Examples include inference engines in a software-defined network, servers in a data center, and drones in a swarm.
In distributed online learning settings, agents take actions over time and receive sequential samples associated with the selected actions. While the agents can cooperate to speed up the learning process, it comes at the expense of communication overhead in sharing sequential samples with others. Hence, distributed online learning problems involve a natural trade-off between learning performance and communication overheads.

\begin{table*}[h]
  \centering
  \footnotesize
  \begin{tabular}{|c|lll|}
    \hline
    \textbf{Algorithm}
     & \textbf{Fully distributed}
     & \textbf{Individual regret}
     & \textbf{Communication cost } \\ \hline
    \texttt{DPE2}~\citep{wang2020optimal}
     & No (leader-follower)
     & \(O(K\log T)\)
     & \(O(K^2M^2\Delta^{-2})\)     \\
    \texttt{ComEx}~\citep{madhushani2021call}
     & Yes
     & \(O(K\log T)\)
     & \(O(KM\log T)\)              \\
    \texttt{GosInE}~\citep{chawla2020gossiping}
     & Yes
     & \(O((K/M + 2)\log T)\)
     & \(\Omega(\log T)\)           \\
    \texttt{Dec\_UCB}~\citep{zhu2021decentralized}
     & Yes
     & \(O((K/M)\log T)\)
     & \(O(MT)\)                    \\
    \texttt{UCB-TCOM}~\citep{wang2023achieve}
     & Yes
     & \(O((K/M)\log T)\)
     & \(O(KM\log\log T)\)          \\\hline\hline
    \balg \textbf{(this work)}
     & Yes
     & \(O((K/M)\log T)\)
     & \(O(KM\log \Delta^{-1})\)    \\
    \hline
  \end{tabular}
  \caption{A comparison summary of prior literature and this work. Note that all algorithms in this table achieve the optimal group regret of \(O(K\log T)\). Hence, we only compare the individual and communication costs of different algorithms.}
  \label{tab:compare-to-prior-literature}
\end{table*}

This paper focuses on studying Cooperative Multi-Agent Multi-Armed Bandit (\CMAB) problems where multiple agents tackle the same instance of a bandit problem. In the standard setting of \CMAB, a set of $M$ independent agents existing over the entire time horizon pull an arm at each time from a common set of $K$ arms. Associated with arms are mutually independent sequences of i.i.d. $[0,1]$-valued rewards with mean $0\le \mu(i)\le 1$, for arm $i\in [K]$.
Each agent has full access to the set of arms: agents are allowed to pull and receive a reward from any arm in the common set without any reward degradation when pulling the same arm.
The goal of each agent is to learn the best arm, with performance characterized by group regret and maximum individual regret according to different application scenarios. In addition to regret, another important metric is the communication overheads that the agents spend in cooperative learning.


The above \CMAB problem is a natural extension of the basic MAB problem \citep{auer2002finite,bubeck2010bandits} in a cooperative multi-agent setting, with extensive recent literature \citep{szorenyi2013gossip,hillel2013distributed,buccapatnam2015information,cesa2016delay,landgren2016distributed,chakraborty2017coordinated,kolla2018collaborative,martinez2019decentralized,sankararaman2019social,feraud2019decentralized,wang2020optimal,dubey2020cooperative,bistritz2020cooperative,chawla2020gossiping,wang2020distributed,madhushani2021one,madhushani2021call,zhu2021decentralized,yang2021cooperative,yang2022distributed,wang2023achieve}.
In terms of solution design, the prior work could be categorized into two paradigms of leader-follower, where a leader agent coordinates the learning process, and fully distributed algorithms, where there is no central coordinator between agents.

In the leader-follower paradigm~\cite{shi2021federatedpersonal,mehrabian2020practical,shi2021heterogeneous, shi2021federated, wang2019distributed, wang2020optimal, bar2019individual, chakraborty2017coordinated, dubey2020cooperative}, a leader agent coordinates the learning process among all agents. The state-of-the-art result in this paradigm is the \texttt{DPE2} algorithm proposed in~\citep{wang2020optimal} and achieves the optimal group regret with a constant number of communication overheads\footnote{Constant communication cost in this paper means it is independent of time horizon $T$.},   Yet, \texttt{DPE2} (and all other leader-follower-based algorithms) relies on a structure where the leader solely pays the exploration costs and incurs almost all the regret in the system. Hence, by nature, this paradigm fails to achieve a good individual regret since all the regret is imposed on the leader agent. It is worth noting that in many practical applications, agents' individual regrets are crucial for a system's overall performance. For example, in a drone swarm, the failure/misbehavior of a single drone, e.g., it crashes into other drones, can dramatically degrade the whole system's overall performance; or in network measurement, the slowest inference engine determines how fast the network parameters, e.g., traffic flows and channel bandwidths, are learned.

An alternative approach is to remove the leader as the central coordinator and design fully distributed cooperative algorithms.
While there has been a success in achieving the optimal group and individual regrets for fully distributed algorithms, they still fail to achieve low communication overhead, such as those in the leader-follower-based algorithms. Early works in this space, e.g.,~\citep{buccapatnam2015information, yang2021cooperative, yang2022distributed} adopted immediate broadcasting as their communication scheme, incurring a high communication cost of $O(T)$. More recent works~\citep{martinez2019decentralized,wang2019distributed,chawla2020gossiping}, improved the communication overhead of the cooperative algorithms to $O(\log T)$ by optimizing the use of communication budget. The state-of-the-art in this line of work is the \texttt{UCB-TCOM} algorithm~\citep{wang2023achieve} that achieves the optimal individual regret of $O(K/M \log T)$ with communication cost of $O(KM \log \log T)$. Despite the above efforts, prior to this work, no existing algorithm, either based on leader-follower or fully distributed, achieves optimal group and individual regret with constant communication costs.

Besides the literature on distributed multi-agent bandits, there is a line of works on batched bandits~\cite{perchet2016batched,gao2019batched,esfandiari2021regret,jin2021almost,karpov2023collaborative} that relate to \CMAB. In batched bandits, the time horizon is separated into several batches, and the reward observations of pulling arms during each batch are only revealed at the end of the batch. This scheme is similar to the distributed bandits, where the observations of other agents after the last communication are only revealed at these agents' next communication. Therefore, the batched bandits algorithm can straightforwardly adapt to our multi-agent bandits setting.
  The current state-of-the-art batched algorithm requires $O(\log \log T)$ batches to attain the near-optimal problem-dependent regret bound~\cite{jin2021almost}. That is, directly transferring their algorithms to the distributed setting leads to $O(\log \log T)$ communication costs. In contrast, our work shows a constant communication cost is enough to guarantee the optimal individual and group regrets.
\subsection{Contributions}\label{subsec:con}

This paper presents \balg, the first fully distributed algorithm that guarantees the optimal group and maximum individual regrets with constant communication costs (see Theorem~\ref{thm:bandit}). Specifically, \balg achieves an $O(\sum_{i:\Delta_i>0}\log T/\Delta_i)$ group regret and an $O((1/M)\sum_{i:\Delta_i>0}\log T/\Delta_i)$ maximum individual regret, where $\Delta_i$ is the gap of reward means between the optimal arm and the $i$-th one. Further, \balg achieves the constant communication complexity of $O(KM\log (1/\Delta))$, where $\Delta = \min_{i}\Delta_i$. A summary of our results and the most relevant prior work is given in Table \ref{tab:compare-to-prior-literature}.

To achieve the above results, \balg leverages a communication policy called Distributed Online Estimation (\alg), which is the main algorithmic contribution of this work. The key idea behind \alg is to determine the synchronization frequency of sharing local empirical estimates of agents so that the quality of estimates is maintained compared to the full cooperation policy. The full cooperation is equivalent to the existence of \textit{centralized estimator} with the best possible estimate using all agents' samples. With limited cooperation, individual agents have access to their locally observed samples, which could cause intrinsic deviations from the centralized estimator.
\alg measures the deviations precisely and uses them as an indicator to trigger a communication round. That is, \alg may urge the agents to communicate with others to synchronize the estimates on the mean of arms once it realizes that the deviation is large. By controlling the deviation in a proper margin, \alg guarantees the optimal learning performance, i.e., the one achievable by the centralized estimator, for individual agents with low communication overheads. By plugging \alg into an elimination-based bandit algorithm, we derive \balg that improves the state-of-the-art results for the \CMAB problem.
Last, using real datasets, we include experiments that demonstrate the improved performance of \balg compared to all benchmark algorithms listed in Table~\ref{tab:compare-to-prior-literature}.
\section{Problem Description} \label{sec:prob}

In the following, we introduce a basic multi-agent multi-armed bandit system model.
We note that the communication policy developed in this paper is generic and could be applied to a broad range of cooperative online learning settings.

Consider a multi-agent stochastic bandit setting with a set $\mathcal{M} = \{1,\dots,M\}$ of independent agents existing over the entire time period, and a set $\mathcal{K}=\{1,2,\dots, K\}$ of arms. Associated with arms are mutually independent sequences of i.i.d. $[0,1]$-valued (e.g., Bernoulli) rewards with mean $0\le \mu(i)\le 1$, for arm $i\in \mathcal{K}$.
Agent $j\in\mathcal{M}$ has full access to the set of arms.
Agents are allowed to pull and receive a reward from any arm from $\mathcal{K}$. Note that for ease of presentation, we focus on a basic model formulation where agents reside on a complete graph, incur no communication delays, and the communication is lossless.
However, the basic model and communication policy proposed in this paper could be extended to account for these practical additions.

In bandit learning, the goal of each agent \(j\) is to learn the best arm as fast as possible with minimizing the \textit{pseudo-regret} (called \textit{regret} for short in the rest of this paper).
The expected regret of an agent $j$ is formally defined as
\begin{equation*}
    \mathbb{E}\left[R_T^j\right] := \mu(i^*) T - \mathbb{E}\left[\sum\nolimits_{t=1}^{T}x_{t}(I_{t}^{j})\right],
\end{equation*}
where $i^*$ is the optimal arm, $I_{t}^{j} $ is the action taken by agent $j$ at round $t$, and $x_{t}(I_{t}^{j})$ is the realized reward. 
Also, the expectation is taken over the randomness of stochastic rewards and the algorithm's (agents') decisions. In a multi-agent setting, the total performance is measured by the total expected regret of all agents, defined as
\begin{equation*}
   \mathbb{E}\left[R_T\right]  :=  \sum\nolimits_{j\in \mathcal{M}}\mathbb{E}\left[R_T^j\right].
\end{equation*}
In addition to the group regret, which characterizes overall performance,
the individual performance of each agent is also important. To capture this individual performance, we measure the maximum individual regret defined as follows,
\begin{equation*}
    \mathbb{E}\left[\bar{R}_T\right]:=\max_{j\in \mathcal{M}}\mathbb{E}\left[R_T^j\right].
\end{equation*}

Similar to other distributed learning problems, the multi-agent MAB setting encourages distributed agents to cooperate with each other by sharing information through \textit{messages}, which include reward observations, reward averages, or arm indices. 
We assume any message can be communicated within a single time slot.  The total number of messages communicated among these agents quantifies the communication complexity of an algorithm.



\section{Algorithm} \label{sec:alg}
This section presents an algorithm that adds a Distributed Online Estimation (\alg) subroutine to each learning agent $j$ and enables them to approximate the estimate of the optimal centralized algorithm having all samples when estimating the parameter of a common i.i.d. process.
We introduce the details of the \alg algorithms in Section~\ref{sec:doe_alg} and then integrate it to a bandit algorithm in Section~\ref{sec:doe_bandit_alg}.

\subsection{Distributed Online Estimation Algorithm (\alg) }
\label{sec:doe_alg}
To facilitate the presentation of the high-level idea of \alg, let us focus on a simplified setting that involves only one arm $i$ whose reward mean $\mu(i)$ is unknown to the distributed agents that sample the process simultaneously in each slot. Since each agent possesses the same number of pulls, we denote $n_t(i)$ as the number of samples available to each agent up to time $t$.
The idea of \alg is to synchronize the estimates of distributed agents when the local estimates deviate substantially from the centralized one with all samples.
By properly configuring \alg, each individual agent manages to efficiently control the deviation of the estimates of individual agents with incurring low communication costs.

More specifically, during the running time, \alg adopts a thresholding policy to decide whether to trigger a communication round so that agents can exchange messages with each other to synchronize their estimates with all samples in the system.
To decide whether to start a communication round, each agent maintains the so-called \textit{Common Mean} (CM) for the mean over all system-wide available samples in the last communication round, and simply compare CM with \textit{Auxiliary Local Estimates} (ALE, details shown in \ref{subsec:ei}).

The value of CM is calculated by averaging all samples in the last communication round, so, its value becomes updated only once at each communication round and remains unchanged in the subsequent non-communication rounds. The value of CM at time $t$ is denoted as $\hat{\mu}_{\texttt{com},t}(i)$. At specific time slots, each agent checks whether the gap between CM and ALE is smaller than some threshold value.
In \alg, all agents share a common threshold value, which can be time-varying with the number of available samples, $n_t(i)$, and thus the threshold value at time $t$ is denoted as $G_{n_t(i)}$.
If the gap between ALE and CM is larger than the threshold value, a new communication round is triggered to synchronize the estimates. By doing so, the sum of new samples from other agents will be collected, a new common mean is calculated, and then the agent broadcasts the new CM to all others.

In \alg, the threshold value $G_{n_t(i)}$ plays a key role in controlling estimate deviations and communication overheads.
Intuitively, when the ALEs of each individual agent center around the common mean, the actual estimates of all agents center around CM as well. Thus, no communication is needed. Otherwise, a communication round is triggered to synchronize the estimates of all agents. Hence, the threshold value determines how far the estimates deviate from each other during the non-communication rounds; the smaller the radius, the smaller the deviations, and the closer the estimates of each agent approach the mean over all samples. On the other hand, with smaller threshold values $G_{n_t(i)}$, agents communicate more frequently with each other. Hence, the trade-off of estimation performance versus communication overheads is associated with $G_{n_t(i)}$.

In the following, we present the technical details of the \alg algorithm and first show how to construct the estimate interval for each agent by using local estimates.

\begin{algorithm}[t]
    \caption{\alg: an algorithm for estimating the mean of arm $i$ by agent $j$, subscript $t$ is dropped}\label{alg:1}
    \footnotesize
    \begin{algorithmic}[1]
        \State \textbf{Parameters:} $\beta>1$; $\ei_n, n=1,2,\dots$
        \State \textbf{Variables:} $\hat{\mu}^j_{\texttt{aux}}(i)$, $n(i)\leftarrow 0$ $\hat{\mu}_{\texttt{com}}(i)\leftarrow 0$, $\ei_{\texttt{last}}\leftarrow G_1$; $X^{j'}(i)\leftarrow 0$, $X_{\texttt{last}}^{j'}(i)\leftarrow 0$, $\forall j'\in \mathcal{M}$
        \For{each round $t$ when the agent gets a new sample}
        \State $n(i)\leftarrow n(i)+1$
        \State Update $X^j(i)$ with the new sample
        \If {$\beta\ei_{n(i)}\leq\ei_{\texttt{last}}$} \label{line:check_s}
        \State $\ei_{\texttt{last}}\leftarrow \ei_{n(i)}$ \label{line:update_g}
        \If {$|\hat{\mu}_{\texttt{aux}}^j(i)-\hat{\mu}_{\texttt{com}}(i)|>\ei_{n(i)}$}\label{line:check_e}
        \State{\texttt{//Communicate to synchronize the estimates}}
        \State{--- Collect $X^{j'}(i)$ from other agents and calculate the new $\hat{\mu}_{\texttt{com}}(i)$} \label{line:communication-1}
        \State --- Broadcast the new $\hat{\mu}_{\texttt{com}}(i)$ to other agents \label{line:communication-2}
        \State --- $X_{\texttt{last}}^{j'}(i)\leftarrow X^{j'}(i)$ for all $j'\in \mathcal{M}$ \label{line:communication-3}
        \EndIf
        \EndIf
        \State Update $\hat{\mu}_{\texttt{aux}}^j(i)$ according to Eq. (\ref{eq:est}) and $\hat{\mu}^j(i)$ according to Eq. (\ref{eq:est_normal})
        \EndFor
    \end{algorithmic}
\end{algorithm}

\subsubsection{Constructing the Auxiliary Local Estimates (ALE)} \label{subsec:ei}

At a non-communication round $t$, an agent only accesses partial external samples from others. Below we introduce how an agent builds up the Auxiliary Local Estimate with missing samples from others.

Note that $n_{t}(i)$ is the number of samples that an agent has made for arm $i$ up to time slot $t$. Let $t_{\texttt{last}}$ denote the last round before $t$ that the condition in Line~\ref{line:check_s} holds,
and $X_{t}^j(i)$ be the sum of rewards from $n_{t}(i)$ samples of agent $j$ at time slot $t$ for arm $i$.
For agent $j$, there are $n_t(i)-n_{t_{\texttt{last}}}(i)$ missing samples from any other agents.
In \alg, agent $j$ uses local samples in the same time slot to compensate the missing samples from other agents to construct ALE, denoted by $\hat{\mu}_{\texttt{aux},t}^j(i)$. That is
\begin{equation}\label{eq:est}
    \hat{\mu}_{\texttt{aux},t}^j(i)=\frac{1}{Mn_t(i)}\sum_{j'=1}^M \left(X_{t_{\texttt{last}}}^{j'}(i)+X_t^j(i)-X_{t_{\texttt{last}}}^j(i)\right)
\end{equation}
where the term $X_t^j(i)-X_{t_{\texttt{last}}}^j(i)$ serves as the compensation for the missing samples from agent $j'$ from $t_{\texttt{last}}$ to $t$. In \alg, ALE mimics the estimate of the estimator, which possesses all $Mn_t(i)$ samples and serves as an index through which the agents decide when to communicate.

We weight the local estimates in ALE such that it may involve a larger estimation error. Hence, in addition to ALE, each agent $j$ calculates the local estimate $\hat{\mu}_{t}^j(i)$ to be used in a bandit algorithm using the following equation.
\begin{equation}\label{eq:est_normal}
    \hat{\mu}_{t}^j(i)=\frac{1}{Mn_{t_{\texttt{last}}}(i)+n_t(i)-n_{t_{\texttt{last}}}(i)}\left(\sum_{j'=1}^M X_{t_{\texttt{last}}}^{j'}(i)+X_t^j(i)-X_{t_{\texttt{last}}}^j(i)\right)
\end{equation}

\subsubsection{Communication Policy of \alg}
Now with the definition of ALE, we present the communication policy of \alg. The pseudocode of \alg is summarized in Algorithm \ref{alg:1}. To decide a communication round, an agent $j$ checks the values of $\hat{\mu}_{\texttt{aux},t}^j(i)$ and $\hat{\mu}_{\texttt{com},t}(i):=(\sum_{j=1}^M X_{t_{\texttt{last}}}^{j}(i))/(Mn_{t_{\texttt{last}}}(i))$ every time the specified threshold value, i.e., $G_{n_t(i)}$, reduces to $1/\beta$ $(\beta> 1)$ times of the original value (Lines \ref{line:check_s}, \ref{line:update_g} in Algorithm \ref{alg:1}). In \alg, $\beta$ determines how frequently the algorithm checks those values.
Once the deviation of the local estimate $\hat{\mu}_{\texttt{aux},t}^j(i)$ from the common mean, $\hat{\mu}_{\texttt{com},t}(i)$, is larger than the specified threshold value, $G_{n_t(i)}$, agent $j$ calls for triggering of a new communication round. In a communication round triggered by agent $j$, the sum of missing samples from the last communication round $t_{\texttt{last}}$ from each other agent will be collected to calculate a new common mean. Then, this new common mean will be broadcast to all other agents.

Our analysis in Lemma \ref{lem:ub} shows that \alg can provide a provable performance guarantee for the single-arm-estimation problem (in the form of confidence interval) with a tunable trade-off between the estimation quality and communication overheads. With a richer communication budget, the estimation performance of \alg approaches that of the optimal estimator with full access to the samples. Since \alg can provide an explicit confidence interval for the mean to be estimated, it is straightforward to plug \alg into bandit algorithms, as exemplified in the next section.

\begin{algorithm}[t]
    \caption{\balg for agent $j$; subscript $t$ is dropped}
    \label{alg:2}
    \footnotesize
    \begin{algorithmic}[1]
        \State \textbf{Parameters:} $\alpha>0$, $\beta> 1$; $G_n$, $n=1,2,\ldots$
        \State \textbf{Initialization:} $\hat{\mu}_{\texttt{com}}^j(i)\leftarrow 0$; $G_{\texttt{last},i}\leftarrow G_1$; $n(i)\leftarrow 0$, $\hat{\mu}^j_{\texttt{aux}}(i)$, for $\forall i$; $G_n\leftarrow \alpha \ci(Mn,\delta)$, $n=1,2,\ldots$
        \For{each round $t$}
        \If {an arm is eliminated by some other agent}
        \State {Update the candidate set}
        \EndIf
        \State Pull an arm $i$ from the candidate set in a round-robin manner \label{line:pull}
        \If{$\beta G_{n(i)}\leq G_{\texttt{last},i}$} \label{line:detection-point}
        \State $G_{\texttt{last},i}\leftarrow G_{n(i)}$
        \If {$|\hat{\mu}^j_{\texttt{aux}}(i)-\hat{\mu}_{\texttt{com}}(i)|>G_{n(i)}$ and the candidate set contains more than one arms}
        \label{line:communication-condition-2}
        \State{Start a communication round to synchronize the estimates on arm $i$}
        \State{\texttt{//Execute Algorithm~\ref{alg:1}'s Lines~\ref{line:communication-1},\ref{line:communication-2},\ref{line:communication-3}}}
        \EndIf\label{line_end}
        \EndIf
        \State Update $n(i)$, $\hat{\mu}^j(i)$ and $\hat{\mu}_{\texttt{aux}}^j(i)$
        \State Update the candidate set via Eq.~\eqref{eq:cand_set}
        \State Notify other agents if an arm is eliminated \label{line:eli}
        \EndFor
    \end{algorithmic}
\end{algorithm}

\subsection{Integrating \alg to a Bandit Learning Algorithm}
\label{sec:doe_bandit_alg}

In this section, we present a distributed bandit algorithm named \balg that uses \alg as the underlying communication policy. We summarize the pseudocode of  \balg in Algorithm \ref{alg:2}.

\balg is based on active arm elimination, which is a classic approach to address the well-known tradeoff between exploration (acquiring new information) and exploitation (optimizing based on available information) in bandit problems. In this approach, the learner constructs a \textit{candidate set} for the arms, which are likely to be optimal, and exploration is allowed only from the arms in the candidate set. When exploring the candidate set, the algorithm periodically pulls an arm in and dynamically eliminates the arms which are unlikely to be optimal.

To integrate \alg with the bandit algorithm, we initiate multiple instances of \alg run by \balg, each of which tackles the estimation of a single arm.
To implement the \alg subroutine, each agent notifies others once an arm is eliminated (Line~\ref{line:eli} in Algorithm \ref{alg:2}) and pulls arms in the candidate set in a round-robin manner (Line~\ref{line:pull}), in order that all agents always pull the same arm at each time slot and \alg is able to keep track of the total number of samples in the system by $Mn_t(i)$ for all agents.
The above rules imply that all agents have a common candidate set, which is denoted by $\mathcal{C}_t$.

\textit{Constructing the candidate set.}
To construct the candidate set, \balg determines an explicit confidence interval for the reward means of arms.  Define $\ci(n,\delta)$ as the radius of the confidence interval for the reward process with $n$ samples and confidence level $1-\delta$. If the reward process is $[0,1]$-valued, we define
\begin{equation}\label{eq:ci_1}
    \ci(n,\delta)=\sqrt{\frac{\log \delta^{-1}}{2n}},
\end{equation}
where $\delta$ specifies the violation probability that the true mean lies outside the above confidence interval.
As we mentioned, the threshold value, $G_{n_t(i)}$, in \alg determines the deviation of the estimates in individual agents from the optimal one with all samples. Hence, in order to guarantee distributed agents to achieve the same order of the convergence rate as the optimal one, we set $G_{n_t(i)}$ according to the confidence interval with the total of $Mn_t(i)$ samples.
By setting $G_{n_t(i)}=\alpha\ci(Mn_t(i),\delta)$ where $\alpha>0$, \alg yields a confidence interval for the mean of arm $i$, whose radius is $(2\alpha\beta+\beta)\ci(Mn_t(i),\delta)$ (see Lemma \ref{lem:ub} on detailed derivation). With the above result, an arm $i$ is eliminated by agent $j$ from the candidate set $\mathcal{C}_t$ at time $t$ if there exist an arm $i' \in \mathcal{C}_t$ such that
\begin{equation}
    \label{eq:cand_set}
    \begin{split}
        \hat{\mu}^j_t(i)+&(2\alpha\beta+\beta)\ci(Mn_t(i),\delta)
        <\hat{\mu}^j_t(i')-(2\alpha\beta+\beta)\ci(Mn_t(i'),\delta).
    \end{split}
\end{equation}

\section{Theoretical Results for Regret and Communication Cost} \label{sec:analysis}

In this section, we summarize the theoretical results, through which we show that the  \balg can achieve the same order of optimal regret as the optimal centralized one while incurring constant communication overheads.

\subsection{Main Results}

The following lemma shows the performance of \alg with the upper bound of estimation error proportional to the radius of the confidence interval with system-wide samples. Then, we summarize the results for \balg in Theorem \ref{thm:bandit}.

\begin{lemma}\label{lem:ub} Assume $M$ agents independently sample an arm with an i.i.d. reward process with unknown mean $\mu(i)$, and $n_t(i)$ is the available samples for each agent up to time slot $t$. With $\beta >1$ and $\ei_{n_t(i)}=\alpha\ci(Mn_t(i),\delta)$, for any $t$, with probability $1-\delta$, we have
    \begin{equation*}
        |\hat{\mu}_t^j(i)-\mu(i)|\leq (2\alpha\beta+\beta)\ci(Mn_t(i),\delta).
    \end{equation*}
\end{lemma}

\begin{theorem}\label{thm:bandit}
    Let $\ci_{[0,1]}(n,\delta)$ in Eq.~\eqref{eq:ci_1} with $1\geq \delta >0$ be the radius of the confidence interval of a $[0,1]$-valued i.i.d. process with $n$ samples. Set $\beta >1$ and $\ei_{n}=\alpha\min\{1, \ci_{[0,1]}( Mn,\delta)\}$, where $\alpha>0$. \emph{\balg} achieves the following performance. \\
    (i) (Group Regret)
    \begin{equation}\label{eq:reg}
        \mathbb{E}\left[R_T\right]\!  =\!O\left(\sum_{i:\Delta_i>0}\!\!\!\frac{8(2\alpha+1)^2\beta^2\log \delta^{-1}}{\Delta_i}\!+\frac{KM^3{\tau}^2T\delta}{2}\!\right)\!,
    \end{equation}
    where $\tau:= \frac{8(2\alpha+1)^2\beta^2\log \delta^{-1}K}{\Delta^2}$.

    (ii) (Maximum individual regret)
    \begin{equation}\label{eq:reg_2}
        \mathbb{E}\left[\bar{R}_T\right]   \!=\!O\left(\sum_{i:\Delta_i>0}\!\frac{8(2\alpha+1)^2\beta^2\log \delta^{-1}}{M\Delta_i}\!+\!\frac{KM^2{\tau}^2T\delta}{2}\right).
    \end{equation}
    (iii) (Communication costs)
    The expected number of messages sent by all agents running \emph{\balg} satisfies the following upper bound.
    \begin{equation}\label{eq:com}
        \sum_{i:\Delta_i>0}6M\log_{\beta}\left(\frac{ 4(2\alpha\beta+\beta)}{ \Delta_i}\right)+\frac{KM^3{\tau}^2T\delta}{2}+M(K-1).
    \end{equation}
\end{theorem}

In what follows, we sketch the high-level idea of the proof of Theorem~\ref{thm:bandit} for both regret and communication costs of \balg. A formal proof is given in Appendix~\ref{app:proof}.
In Section \ref{sec:alg}, we tailor the elimination-based strategy in \balg such that all agents pull arms in a synchronized manner. Hence, each agent can always track the number of pulls of any arm $i$ by $Mn_t(i)$. In this way, \CMAB involves multiple distributed online estimation problems, each of which can be solved by \alg separately.
In Lemma \ref{lem:ub}, we show that the \alg subroutine builds up a confidence interval for the mean reward of an arm (involving an additional constant factor compared to the one in single-agent settings with all samples always being available).
Then, we can prove the regret bound in Theorem \ref{thm:bandit} by applying the results of Lemma~\ref{lem:ub} to the standard analysis for a multi-armed bandit problem.  Specifically, the expected number of pulls when the true mean of an arm is outside of the characterized confidence interval is very small. Thus the probability of eliminating the optimal by \balg is low. In other words, the major part of the regret of \balg is introduced during the elimination phase when the candidate set contains more than one arm. Upper bounding the length of the above-mentioned elimination phase of \balg yields the regret result in Eq.~\eqref{eq:reg_2}.

To prove the communication cost, we highlight the fact that agents communicate the mean of arm $i$ only when the radius of the confidence interval provided by \alg is larger than $\Delta_i/2$ (such that the investigated arm remains in the candidate set). Combining with the rule of \alg that agents communicate only when the radius of the confidence interval reduces to $1/\beta$ of the previous, we prove the bound for the communication cost.

\subsection{Discussion}
In the following, we discuss several remarks regarding the significance of our results.

\paragraph{Optimality.}
When $\delta= O(1/T^s)$, $s\geq 2$, the second term in Eq. (\ref{eq:reg}) and (\ref{eq:reg_2}) becomes constant. Hence, we can recover a $O(\sum_{i:\Delta_i>0}(1/\Delta_i)\log T)$ group regret and $O(\sum_{i:\Delta_i>0}(1/\Delta_i)\log T/M)$  individual regret for the distributed bandit problem, implying that the proposed algorithm attains both the (order-) optimal group and maximum individual regrets. In the meantime, it is possible to drop the second term in Eq. (\ref{eq:com}), and thus
\balg incurs constant $O(MK\log (1/\Delta))$ communication costs. This result substantially improves the state-of-the-art result for fully distributed bandit algorithms (see Table \ref{tab:compare-to-prior-literature}).

\paragraph{Influence of $\alpha$ and $\beta$.}
Eq.~\eqref{eq:com} shows that communication overheads influence the estimation quality through parameters $\alpha$ and $\beta$. Generally speaking, $\beta$ specifies the frequency that \alg checks the deviation of individual estimates, directly upper bounding the communication overheads for \balg. Hence, $\beta$ seems to have a larger influence in the communication overheads bound in Theorem \ref{thm:bandit} than $\alpha$. On the other hand, $\alpha$ specifies the radius of the estimate interval or the threshold for the estimate deviation, which triggers an actual communication demand. Thus, the influence of $\alpha$ is more reflected in the empirical performance.
Actually, the empirical performance of \balg in communication complexity can be much better than the theoretical bound since agents running the \balg algorithm start a communication round in an on-demand manner, i.e., only when their estimates deviate a lot from each other. For example, if the investigated process is benign, our algorithm can achieve much lower communication overheads empirically than those works whose communication policies fail to adapt to dynamic environments.

\paragraph{Results for other i.i.d. processes.} \balg triggers a communication round based on the variation of the threshold, with the communication overheads on a suboptimal arm $i$ being $O(\log_{\beta}(\ei_{1}/\ei_{n_T(i)}))$, approximately. In \balg, the threshold value is set based on that of the confidence interval with all samples (up to a tunable parameter $\alpha$).
For a Bernoulli process, the mean always lies in $[0,1]$. Hence, we can set $\ei_1=1$, which results in $O(\log(1/\ei_{n_T}(i)))$ or $O(\log(1/\Delta))$ communication overheads.
We note that by slight modification, the \balg algorithm can tackle other i.i.d. processes with similar results obtained.
For an i.i.d. process with an unbounded mean, such as the Gaussian process, the \balg may choose to start a communication round only when the size of the confidence interval shrinks to
$O(\sqrt{M})$. This will not degrade the regret results guaranteed in Theorem \ref{thm:bandit}, since the algorithm only has to spend on average $O(\log T)$ samples in shrinking the confidence intervals of all arms, with an increase of $O(K\log T)$ regret. On the other hand, the communication overheads is only $O(\log(\sqrt{M}/\Delta))$, since $\ei_1$ can be set to $O(\sqrt{M})$.

\begin{figure*}[t]
    \centering
    \subfloat{\includegraphics[width=0.7\textwidth]{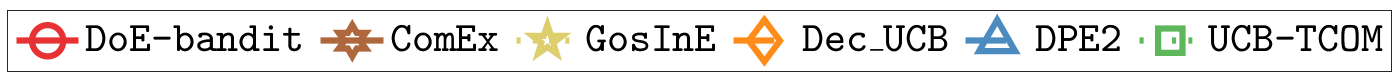}}\\
    \setcounter{subfigure}{0}
    \vspace{-10pt}
    \subfloat[Communication cost]{\includegraphics[width=0.32\textwidth]{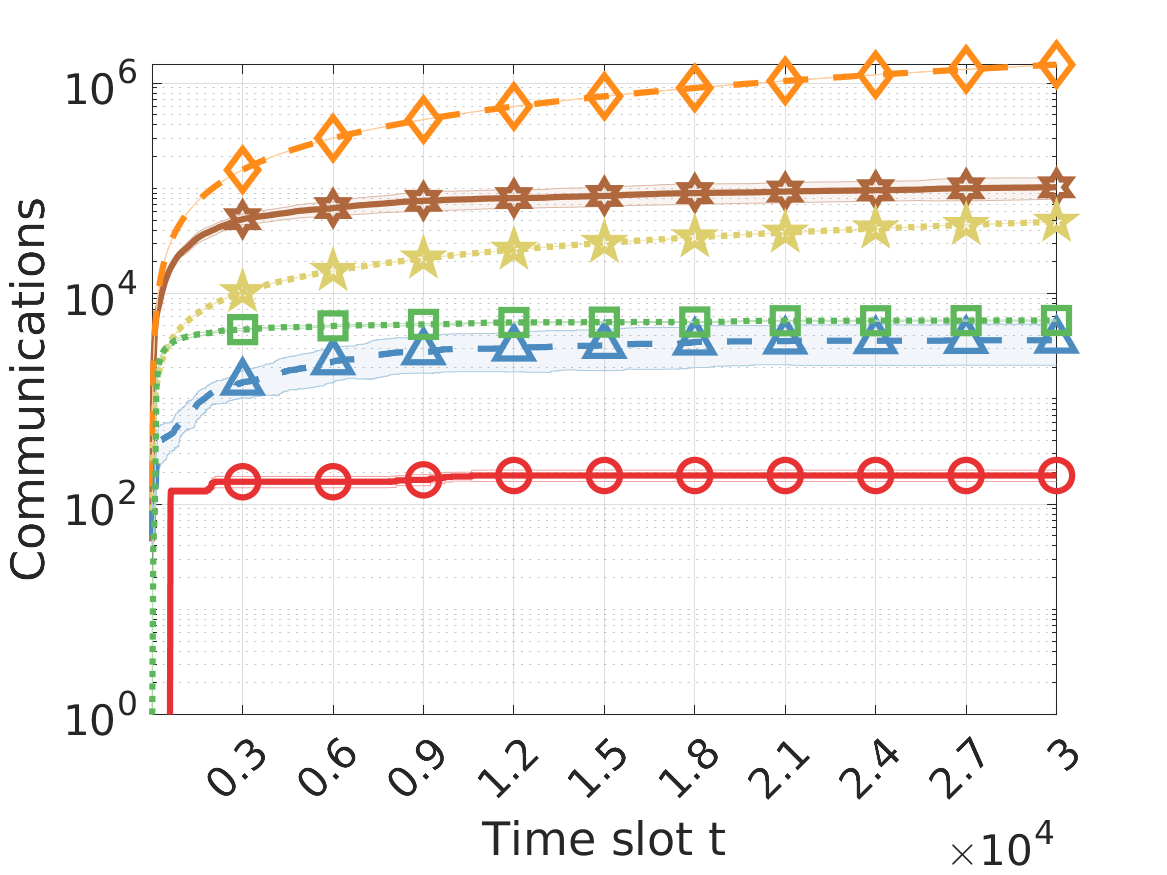}
        \label{subfig:communication}}
    \hfill
    \subfloat[Group regret]{\includegraphics[width=0.32\textwidth]{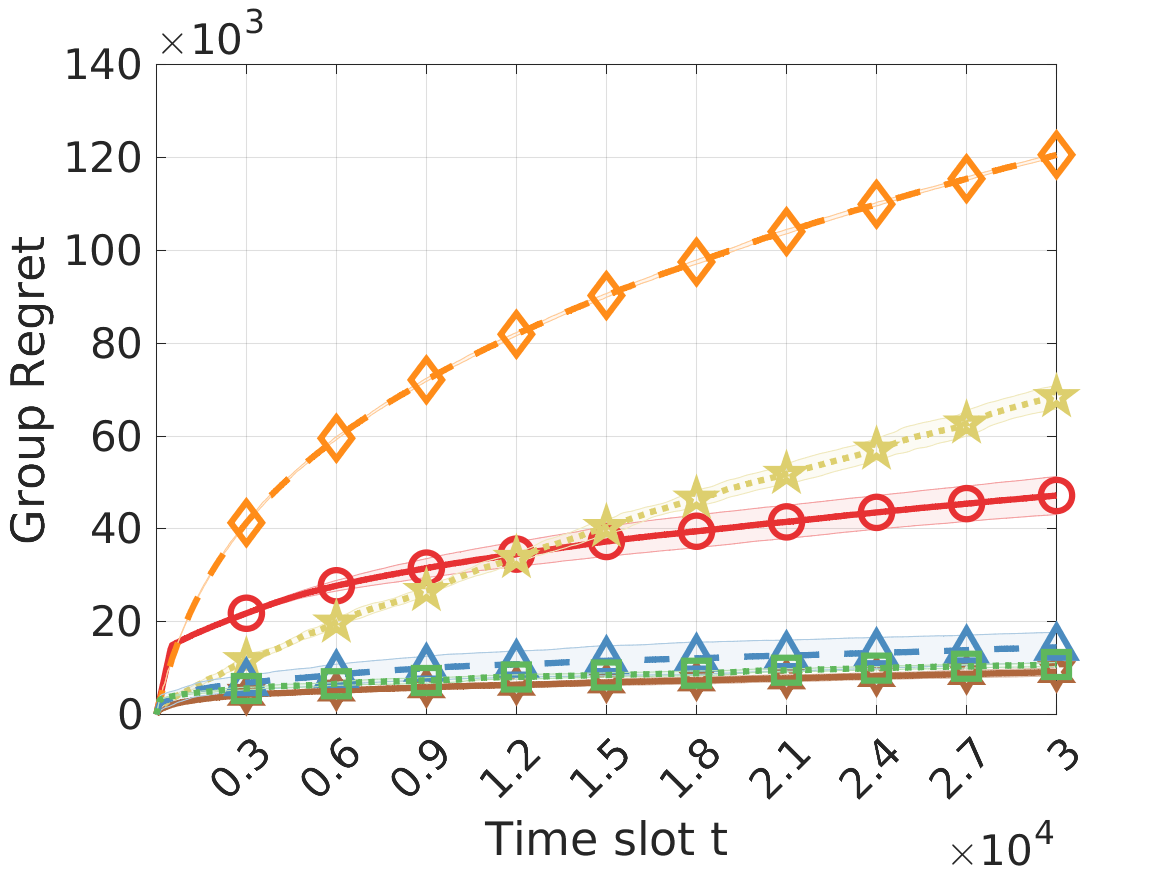}
        \label{subfig:group-regret}}
    \hfill
    \subfloat[Individual regret]{\includegraphics[width=0.32\textwidth]{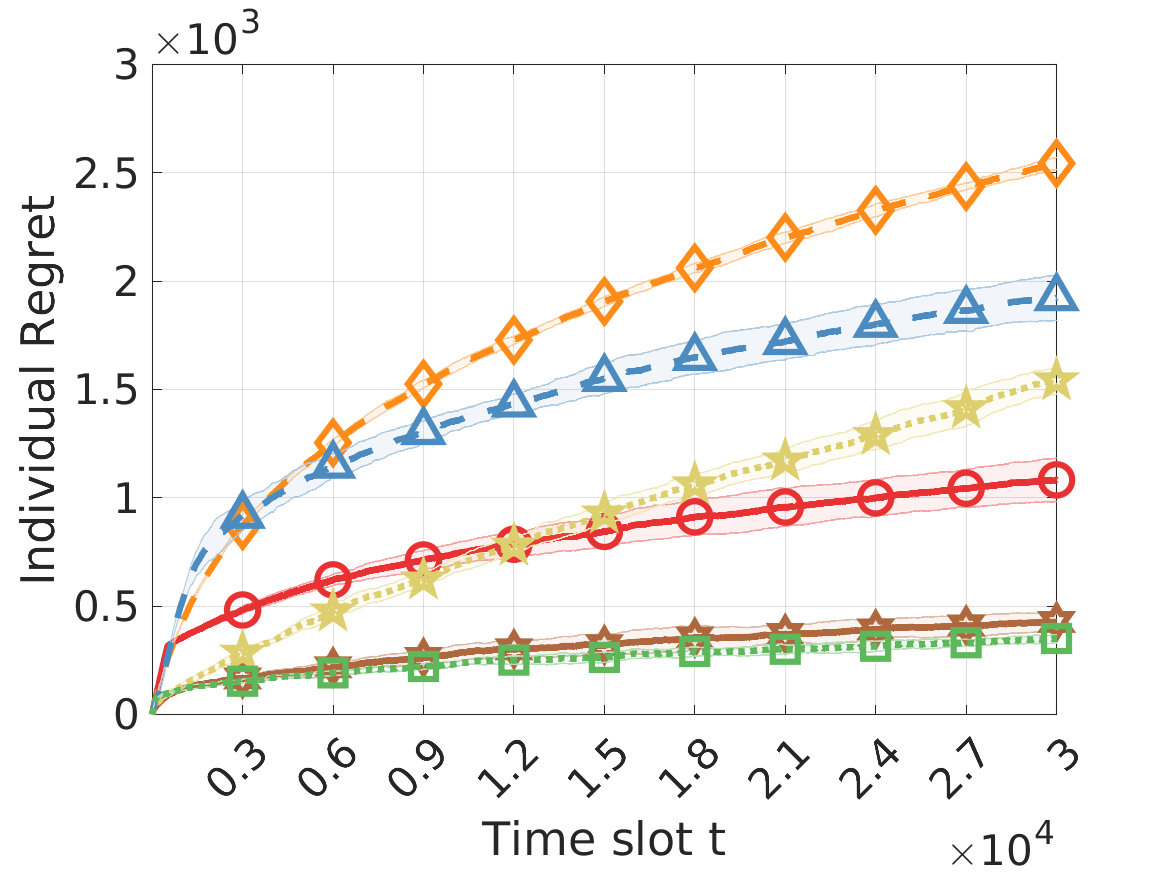}
        \label{subfig:individual-regret}}
    \hfill
    \vspace{-2mm}
    \caption{\balg (this work) vs. baseline algorithms listed in Table~\ref{tab:compare-to-prior-literature}}
    \label{fig:baseline-comparison}
\end{figure*}

\begin{figure*}[t]
    \centering
    \hfill
    \subfloat[Vary reward gap (\(10\) arms, \(5\) agents)]{\includegraphics[width=0.32\textwidth]{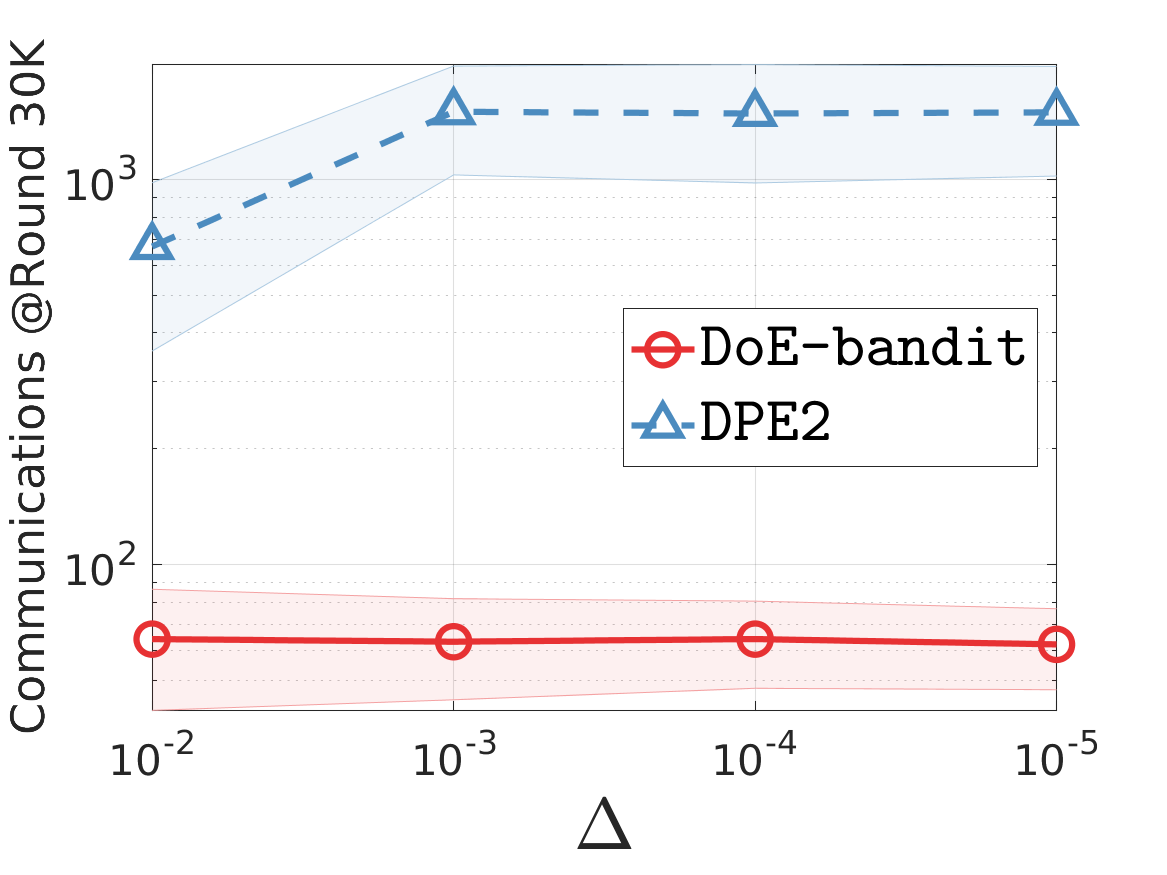}
        \label{subfig:comm-delta}}
    \hfill
    \subfloat[Vary agent number (\(20\) arms)]{\includegraphics[width=0.32\textwidth]{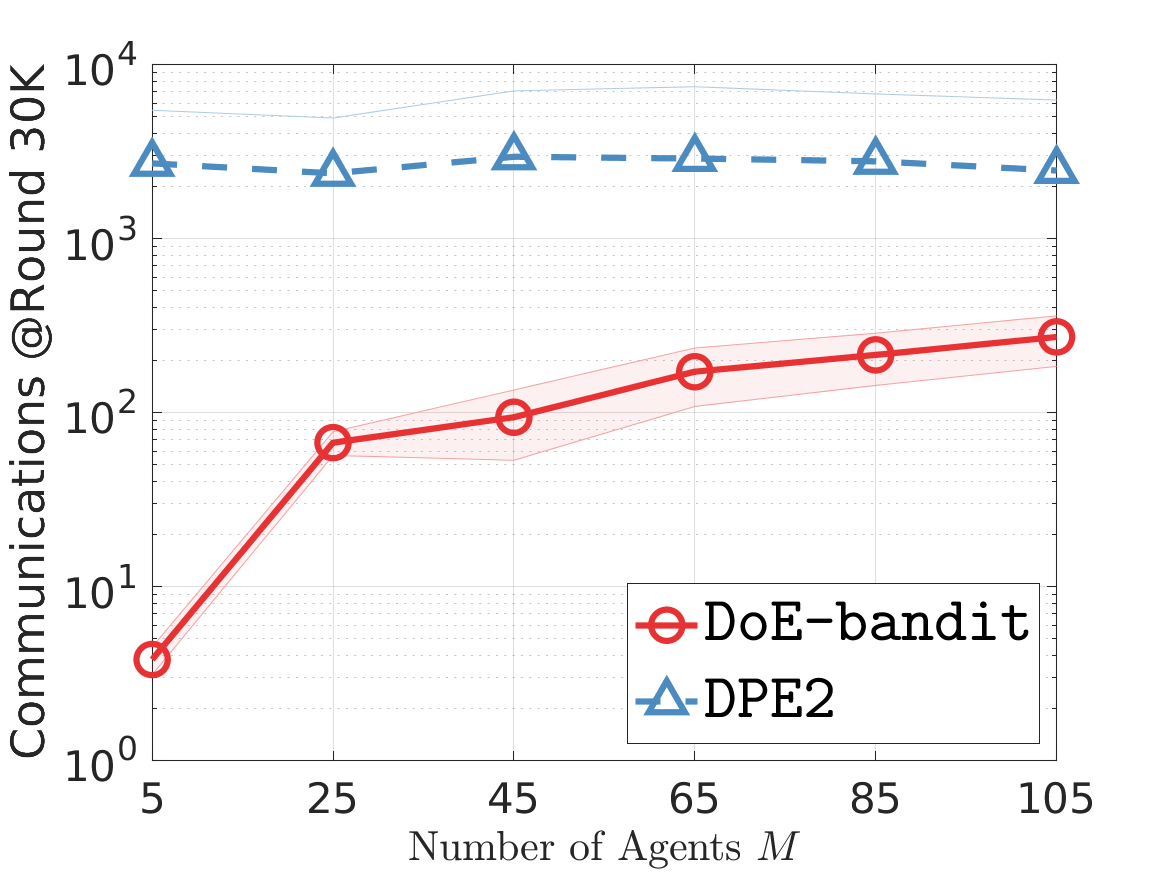}
        \label{subfig:comm-agent}}
    \hfill
    \subfloat[Vary arm number (\(25\) agents)]{\includegraphics[width=0.32\textwidth]{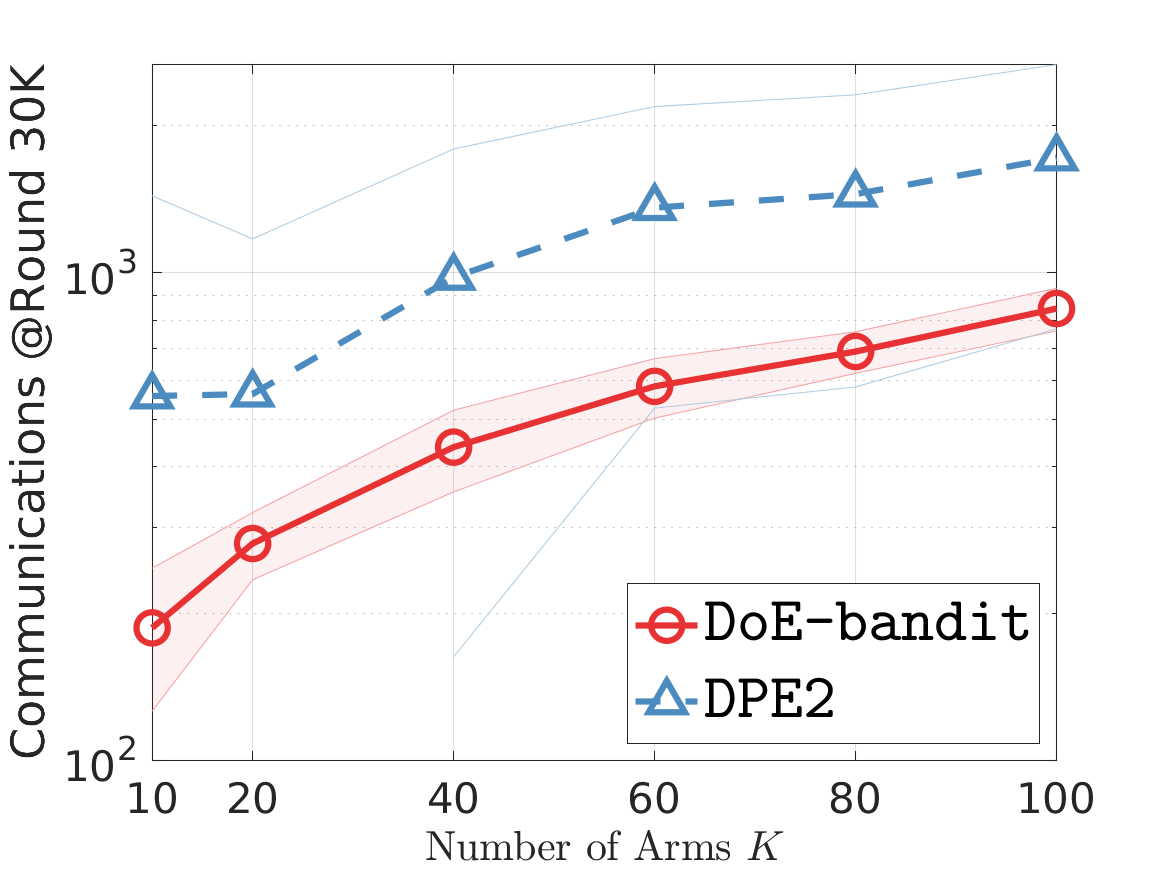}
        \label{subfig:comm-arm}}
    \hfill
    \vspace{-2mm}
    \caption{
        Communications: \balg vs. \texttt{DPE2}
    }
    \label{fig:communication}
\end{figure*}
\section{Numerical Results} \label{sec:sim}

In this section, we conduct numerical experiments to corroborate the performance of the \balg algorithm. We aim to highlight the advantage of \balg in group and individual regrets and in communication costs over start-of-the-art baselines.

\paragraph{Experimental Setups and Baseline Algorithms}
We consider a multi-agent bandits setting with $K=100$ arms, $M=50$ agents, and \(T=30K\), and each arm is associated with a Bernoulli distribution with mean randomly taken from the click-through-rate in \texttt{Ad-Clicks}~\cite{adclicks}.
In \texttt{DoE-bandit} algorithm, we set parameters \(\alpha=1,\beta=3\) and \(\delta=1/T^2\).
We run \(50\) trials of  each experiment and plot the means as lines and their standard deviations as shaded regions.

We compare the regret and communication costs of \balg with five baselines (\texttt{ComEx}~\citep{madhushani2021call},
\texttt{GosInE}~\citep{chawla2020gossiping},
\texttt{Dec\_UCB}~\citep{zhu2021decentralized},
\texttt{DPE2}~\citep{wang2020optimal},
and \texttt{UCB-TCOM}~\citep{wang2023achieve}) outlined in Table~\ref{tab:compare-to-prior-literature}. We note that some of the baseline algorithms are developed for a set of agents that are connected through an underlying graph topology. Hence, to make the comparison fair, we consider a complete graph for all algorithms so that any two agents can communicate.

\paragraph{Experimental Results}
Figure~\ref{fig:baseline-comparison} reports the comparison results. Figure~\ref{subfig:communication} shows that \balg achieves the smallest communication costs among all algorithms. Note that \balg and \texttt{DPE2} are the only two algorithms with constant communication costs, better than others and matching the theoretical results in Table \ref{tab:compare-to-prior-literature}.
Figure~\ref{subfig:group-regret} reports the group regrets of algorithms. The results show \balg is not as good as \texttt{DPE2}, \texttt{ComEx}, and \texttt{UCB-TCOM}.
This is because \balg is based on the arm-elimination policy and others are UCB-like algorithms. It is known that with the same order-wise regret performance, UCB algorithms are empirically better than elimination ones in general~\citep[\S6]{garivier2016explore}.
Figure~\ref{subfig:individual-regret} reports the maximum individual regrets of agents. UCB-like algorithms perform still better than others. However, \texttt{DPE2}---the other algorithm with constant communication cost---suffers poor individual regret since it leverages a leader-follower structure to complete the learning task, and the leader agent incurs high individual regret in the leader.

To further investigate the communication costs of \balg, in Figure~\ref{fig:communication}, we report the communication costs of \balg in comparison to \texttt{DPE2}, as the only alternative with constant communication costs, under a variety of different parameter settings.
We study the impact of three parameters on the communications costs of \balg and \texttt{DPE2}: (1)
the reward gap \(\Delta\) between arms in Figure~\ref{subfig:comm-delta}; (2)
the number of agents \(M\) in Figure~\ref{subfig:comm-agent};
and (3) the number of arms \(K\) in Figure~\ref{subfig:comm-arm}.
The log-y-axis of these three figures is the final cumulative communication costs at the end of the time horizon.
In all figures, the communication cost of \balg is always better than \texttt{DPE2}'s.
Figure~\ref{subfig:comm-delta} shows when \(\Delta\) decreases, the communication costs of \balg only change slightly while \texttt{DPE2}'s increase.
This is because the communication cost of \balg is \(O(KM\log\Delta^{-1})\) which is much better than that of \texttt{DPE2} which is \(O(K^2M^2\Delta^{-2})\).
Last, the communication costs of both \balg and \texttt{DPE2} increases as the number of agents \(M\) (Figure~\ref{subfig:comm-agent}) or arms \(K\) (Figure~\ref{subfig:comm-arm}) increases.
This corroborates their communication cost upper bounds' dependency on \(K\) and \(M\).


\section{Conclusions}
This paper presented \balg, a fully distributed algorithm for a cooperative multi-agent multi-armed bandits problem. The proposed algorithm achieves the optimal group and individual regret with constant communication overhead. The theoretical claims are verified by numerical experiments and show that \balg outperforms prior algorithms.

The core communication policy proposed in this paper could be further extended in multiple directions. To address the exploitation-exploration dilemma in bandit learning, \balg adopts an elimination-based strategy to determine the arms which will be pulled. The elimination-based strategy is thought to be less practically efficient than others, such as the UCB strategy (Upper Confidence Bound) and the TS strategy (Thompson sampling). This phenomenon is also observed in the multi-agent multi-armed bandit setting (see our experimental results in Figure~\ref{fig:baseline-comparison}).
Hence, it is meaningful to develop an UCB/TS-based algorithm which achieves better practical performance with guaranteeing the same optimal theoretical results claimed in this work.
Second, one can extend the work to capture more practical concerns, such as considering an underlying topology for agents, communication delays between agents, and lossy communication between agents.



\newpage




\bibliographystyle{plainnat}
\bibliography{ref}

\newpage
\appendix
\onecolumn
\newpage
\appendix
\label{sec:app}

\section{Proofs}
\label{app:proof}
\subsection{Proof for Main Theorem \ref{thm:bandit}}

\paragraph{Proof for the Regret Results}
By running the \alg subroutine, the bandit learning algorithm can build up a confidence interval for the mean reward of an arm. In Lemma \ref{lem:ub}, we provide the estimation performance of \alg in estimating the mean of an arm.

According to the results in Lemma \ref{lem:ub}, letting $\ei_{n}=\alpha\min\{ 1,\ci_{[0,1]}( Mn,\delta)\}$, each agent can attain the order-optimal estimate (up to a constant factor $2\alpha\beta+\beta$) for the mean reward, which slightly degrades the performance of the bandit algorithm. We prove the regret of \balg by using the observation in Lemma~\ref{lem:ub}.

In our analysis, we categorize decisions made by the agents into  Type-\uppercase\expandafter{\romannumeral1} and Type-\uppercase\expandafter{\romannumeral2} decisions. Type-\uppercase\expandafter{\romannumeral1}  corresponds to the decisions of an agent when the true mean values of all arms lie in the confidence intervals calculated by each agent, i.e., for any arm \(i\) and agent \(j\),
\begin{equation*}
    \begin{split}
        \mu \in \left[ \hat{\mu}^j_t\left(i\right) - (2\alpha\beta+\beta)\ci_{[0,1]}(Mn_t(i),\delta),\hat{\mu}^j_t\left(i\right) + (2\alpha\beta+\beta)\ci_{[0,1]}(Mn_t(i),\delta)\right].
    \end{split}
\end{equation*}
Otherwise, Type-\uppercase\expandafter{\romannumeral2} decision occurs, i.e., the actual mean value of some arm is not within the confidence interval calculated by some agents.  Note that agents may incur high regret when wrongly eliminating the optimal arm from the candidate set at some time slot with making a Type-\uppercase\expandafter{\romannumeral2} decision.
To prove the regret, we upper bound the probability that a Type-\uppercase\expandafter{\romannumeral2} decision happens and the number of pulls of suboptimal arms without any  Type-\uppercase\expandafter{\romannumeral2} decision occurring, respectively.

We first upper bound the probability that there is a Type-\uppercase\expandafter{\romannumeral2} decision.
Note that an agent makes a Type-\uppercase\expandafter{\romannumeral2} decision once the true mean of some arm is outside the confidence interval. For any time slot $t$, any agent $j$ and any arm $i$, we have
\begin{equation*}
    \begin{split}
        &\Pr\left[\mu(i) \notin \left[ \hat{\mu}^j_t\left(i\right) - F(n_t(i)),\hat{\mu}^j_t\left(i\right) + F(n_t(i))\right]\right] \\
        &\qquad\qquad\qquad\qquad = \sum_{n=1}^{Mt}\Pr\left.\left[\mu(i)\! \notin \left[ \hat{\mu}^j_t\left(i\right) - F(n),\hat{\mu}^j_t\left(i\right) + F(n)\right]\right|n_t(i)=n\right]
        \leq Mt\delta,
    \end{split}
\end{equation*}
where $F(n_t(i)):=(2\alpha\beta+\beta)\ci_{[0,1]}(Mn_t(i),\delta)$.
Then, an agent at any time slot makes a Type-\uppercase\expandafter{\romannumeral2} decision with probability at most $KMt\delta$ (there are $K$ arms).
Hence, with a union bound, the probability that a Type-\uppercase\expandafter{\romannumeral2} decision has happened before a time slot $s$ can be obtained by summing up the above probabilities over investigated time slots (up to \(s\)) and agents, which is
$KM^2s^2\delta/2$.

Now we proceed to upper bound the number of pulls of a suboptimal arm  with only Type-\uppercase\expandafter{\romannumeral1} decisions happening.
According to the rule of the elimination-based bandit algorithm, a suboptimal arm will be removed from the candidate set without further consideration only when the radius of the confidence interval for this arm reduces to a small value with enough samples.
By the following lemma, we upper bound the number of pulls of suboptimal arms by agents ($Mn_t(i)$) when Type-\uppercase\expandafter{\romannumeral1} decision happens.

\begin{lemma}\label{lem:ub_1}
    At any time $t\leq T$, if the optimal arm lies in the candidate set and an agent makes a Type-\uppercase\expandafter{\romannumeral1} decision with pulling a suboptimal arm $i$, i.e., $I_t^j=i$, there is
    \begin{equation*}
        M n_t(i)\leq \frac{8(2\alpha+1)^2\beta^2\log \delta^{-1}}{\Delta^2_i}+M.
    \end{equation*}
\end{lemma}

Lemma \ref{lem:ub_1} holds for any arm, and, therefore,
the total number of times of pulling all $K$ arms (before elimination, in a round-robin manner) is upper bounded by $\tau \coloneqq \frac{8(2\alpha+1)^2\beta^2\log \delta^{-1}K}{\Delta^2}$.
Hence, if there is no Type-\uppercase\expandafter{\romannumeral2} decision happening before $\tau$, the optimal arm will stay in the candidate set all the time, and the regret of \balg in this case is
\begin{equation*}
    \begin{split}
        \sum_{i:\Delta_i>0}M n_t(i)\Delta_i
        =\sum_{i:\Delta_i>0}\left(\frac{8(2\alpha\beta+\beta)^2\log \delta^{-1}}{\Delta_i}+M\Delta_i\right).
    \end{split}
\end{equation*}
On the other hand, if there is a Type-\uppercase\expandafter{\romannumeral2} decision happening before $\tau$, we can upper bound the regret of \balg by $MT$.

Last, with the regrets in the cases with/without a Type-\uppercase\expandafter{\romannumeral2} decision, we can upper bound the expected regret of the \balg algorithm.
\begin{equation*}
    \begin{split}
        \mathbb{E}\left[R_T\right] \leq & \sum_{i:\Delta_i>0}\left(\frac{8(2\alpha\beta+\beta)^2\log \delta^{-1}}{\Delta_i}+M\Delta_i\right)+\frac{KM^3{\tau}^2T\delta}{2},\\
    \end{split}
\end{equation*}
where the first term on the right-hand side corresponds to the regret portion when there is no Type-\uppercase\expandafter{\romannumeral2} decision before $\tau$, and the second term corresponds to the other case.

Due to the same round-robin arm pulling manner in Line~\ref{line:pull} of Algorithm~\ref{alg:2}, all agents by \balg pull the same arm at any time slot, and, therefore, all agents' individual regrets (rewards) are equal. So, we obtain the individual regret for each agent by dividing the above total regret upper bound equally.
We summarize the above results and give the regret upper bounds for \balg in Theorem \ref{thm:bandit}.

\paragraph{Proof for the Communication Costs}
We analyze the communication overheads of \balg arm by arm. If there is a Type-\uppercase\expandafter{\romannumeral2} decision before $\tau$, we use $MT$ to upper bound the communication overheads. The expected communication complexity in this case is then
\begin{equation}\label{eq:ub_5}
    \frac{KM^3{\tau}^2T\delta}{2}.
\end{equation}
In the following, we focus on Type-\uppercase\expandafter{\romannumeral1} decisions.
For any suboptimal arm \(i\) (with \(\Delta_i > 0\)),
let $\tau_i$ be the last time that \balg pulls the arm $i$. At $\tau_i$, we have
\begin{equation*}
    4(2\alpha\beta+\beta)\ci_{[0,1]}\left(Mn_{\tau_i}(i),\delta\right) \geq\Delta_i.
\end{equation*}
The above equation is proved in the proof of Lemma \ref{lem:ub_1} (at Appendix~\ref{subsec:proof-of-lemma-ub_1}). When setting $\ei_{n}=\alpha\min\{1, \ci_{[0,1]}( Mn,\delta)\}$, there is $\ei_{n_{\tau_i}(i)}=\alpha\ci_{[0,1]}( Mn_{\tau_i}(i),\delta)$, and
\begin{equation*}
    4(2\alpha\beta+\beta)\frac{1}{\alpha}\ei_{n_{\tau_i}(i)}\geq\Delta_i.
\end{equation*}

Hence, up to time $\tau_i$, the communications due to arm $i$ is (recall \(G_1 = \alpha\))
\begin{equation*}
    \log_{\beta}(\ei_1/\ei_{n_{\tau_i}(i)})\leq \log_{\beta}\left(\frac{ 4(2\alpha\beta+\beta)}{ \Delta_i}\right).
\end{equation*}

The expected number of communications by suboptimal arms is at most
\begin{equation*}
    \sum_{i:\Delta_i > 0}\log_{\beta}\left(\frac{ 4(2\alpha\beta+\beta)}{\alpha \Delta_i}\right).
\end{equation*}

For the optimal arm, the number of communications (when there is no Type-II decision) can be upper bounded by the largest communication overheads of suboptimal arms.
That is, the number of communications about the optimal arm is upper bounded by $O(\log_{\beta}(1/\Delta))$ where the $\Delta$ corresponds to the smallest non-zero reward gap.
That is because when there are multiple arms in the candidate set, the optimal arm with others in the candidate set is pulled in a round-robin manner and incurs the same communication overheads as others in the set; and when there is only one arm left in the candidate set, the \balg stops communication.
So, to sum up, the total communication overheads is upper bounded by \begin{equation*}
    \sum_{i:\Delta_i > 0}\log_{\beta}\left(\frac{ 4(2\alpha\beta+\beta)}{ \Delta_i}\right)
    + \log_{\beta}\left(\frac{ 4(2\alpha\beta+\beta)}{ \Delta_{\text{min}}}\right)
    \le 2\cdot \sum_{i:\Delta_i > 0}\log_{\beta}\left(\frac{ 4(2\alpha\beta+\beta)}{ \Delta_i}\right)
\end{equation*}

At each communication time, agents spend totally $3M$ messages in collecting messages and synchronize the estimates in each agent. In addition, \alg may update the candidate set in agents when an arm is eliminated, that costs another $M(K-1)$ messages. Therefore, combined with Eq. (\ref{eq:ub_5}), the expected communication overheads of \balg (the total number of messages) is upper bounded by Eq. (\ref{eq:com}).

\subsection{A Proof of Lemma \ref{lem:ub}}

We prove the lemma by analyzing the following two cases.
Let \(s\) denote the last detection point, i.e., the last time slot (before \(t\)) that the condition in Line~\ref{line:detection-point} of Algorithm~\ref{alg:2} holds.

Case (1): the agent communicated at the last detection point $s$. In this case, the estimate $\hat{\mu}_{t}^j(i)$ is obtained by averaging $Mn_s(i)+n_t(i)-n_s(i)$ samples. Hence, the following equation holds with probability $1-\delta$.
\begin{equation*}
    \begin{split}
        |\hat{\mu}_{t}^j(i)-\mu(i)|\leq &\ci_{[0,1]}(Mn_s(i)+n_t(i)-n_s(i),\delta) \\
        \overset{(a)}\leq & \ci_{[0,1]}(Mn_s(i),\delta)\\
        \overset{(b)}\leq & \beta \ci_{[0,1]}(Mn_t(i),\delta),
    \end{split}
\end{equation*}
where the inequality (a) is due to that the width of confidence interval increases with smaller number of samples, and the inequality (b) is due to that the condition in Line~\ref{line:detection-point} is false at time slot \(t\).

In this case, the result in Lemma~\ref{lem:ub} holds.

Case (2): there is no communication at $s$.
Let $A$ be the sum of samples obtained by agent $j$ if communication \emph{happened} at $s$.
We have
\begin{equation*}
    \begin{split}
        &\left|(Mn_s(i)+n_{t}(i)-n_s(i))\hat{\mu}_{t}^j(i)-A\right| \\
        =&\left|\left(Mn_s(i)\hat{\mu}_{\texttt{aux},s}^j(i) + \left(X_t^j(i)-X_s^j(i)\right)\right)
        -\left(\sum_{j'=1}^M X_s^{j'}(i) + \left(X_t^j(i)-X_s^j(i)\right)\right) \right| \\
        =&\left|Mn_s(i)\hat{\mu}_{\texttt{aux},s}^j(i)-\sum_{j'=1}^M X_s^{j'}(i)\right|.
    \end{split}
\end{equation*}

The above equation is based on the fact that agent always has the local samples after $s$ no matter there is communication at $s$.

Hence,
\begin{equation*}
 \begin{split}
        \left|\hat{\mu}_{t}^j(i)-\frac{A}{Mn_s(i)+n_{t}(i)-n_s(i)}\right|
        &=\frac{1}{(Mn_s(i)+n_{t}(i)-n_s(i))}\left|Mn_s(i)\hat{\mu}_{\texttt{aux},s}^j(i)-\sum_{j'=1}^M X_s^{j'}(i)\right|\\
         &=\frac{Mn_s(i)}{(Mn_s(i)+n_{t}(i)-n_s(i))}\left|\hat{\mu}_{\texttt{aux},s}^j(i)-\frac{1}{Mn_s(i)}\sum_{j'=1}^M X_s^{j'}(i)\right|\\
        &\leq   \left|\hat{\mu}_{\texttt{aux},s}^j(i)-\frac{1}{Mn_s(i)}\sum_{j'=1}^M X_s^{j'}(i)\right|   \\
        &\leq  2\alpha \ci_{[0,1]}(Mn_s(i),\delta)\leq 2\alpha\beta \ci_{[0,1]}(Mn_{t}(i),\delta).
    \end{split}
\end{equation*}
where the second inequality is because the condition in Line~\ref{line:communication-condition-2} in Algorithm~\ref{alg:2} does not hold at time slot \(s\) (since there is no communication at \(s\)).
In this case, $ \left|\hat{\mu}_{\texttt{aux},s}^j(i)-\hat{\mu}_{\texttt{aux},s}^{j'}(i)\right|\leq 2\ei_{n_s(i)} =   2\alpha \ci_{[0,1]}(Mn_s(i),\delta)$ for any agents $j$ and $j'$.
Also, $\frac{1}{Mn_s(i)}\sum_{j'=1}^M X_s^{j'}(i)$ averages all samples up to $s$, and hence its value lies between $\min_{j'}\hat{\mu}_{\texttt{aux},s}^{j'}(i)$ and $\max_{j'}\hat{\mu}_{\texttt{aux},s}^{j'}(i)$, which weight partial local samples with a factor $M$ to replace missing ones. Combining the two facts, we prove the second inequality.

Since $A$ contains the same set of samples as the \(\hat{\mu}_{t}^j(i)\) in Case (1), the following equation also holds with probability $1-\delta$:
\begin{equation*}
    \left|\frac{A}{Mn_s(i)+n_{t}(i)-n_s(i)}-\mu(i)\right|\leq\beta \ci_{[0,1]}(Mn_t(i),\delta).
\end{equation*}

Combining the above two equations yields
\begin{equation*}
    \left|\hat{\mu}_{t}^j(i)-\mu(i)\right|\leq(2\alpha+1) \beta\ci_{[0,1]}(Mn_t(i),\delta), ~\text{with ~probability}~1-\delta.
\end{equation*}

As a result, we prove the  Lemma~\ref{lem:ub}.

\subsection{A Proof of Lemma \ref{lem:ub_1}}\label{subsec:proof-of-lemma-ub_1}
We consider agent $j$ running the proposed algorithm makes a Type-\uppercase\expandafter{\romannumeral1} decision at time $t$ and $I_t^j=i$. First, we claim that the following holds.
\begin{equation} \label{eq:con_2}
    2(2\alpha\beta+\beta)\ci_{[0,1]}\left(Mn_{t}(i),\delta\right)+2(2\alpha\beta+\beta) \ci_{[0,1]}\left(Mn_{t}(i^*),\delta\right)\geq\Delta_i.
\end{equation}

Otherwise, we have
\begin{equation*}
    \begin{split}
        &\quad\hat{\mu}^j_t\left(i^*\right)-(2\alpha\beta+\beta)\ci_{[0,1]}\left(Mn_{t}(i^*),\delta\right)\\
        &=\hat{\mu}^j_t\left(i^*\right)+(2\alpha\beta+\beta)\ci_{[0,1]}\left(Mn_{t}(i^*),\delta\right)-2(2\alpha\beta+\beta)\ci_{[0,1]}\left(Mn_{t}(i^*),\delta\right)\\
        &\geq\mu\left(i^*\right)-2(2\alpha\beta+\beta)\ci_{[0,1]}\left(Mn_{t}(i^*),\delta\right)\\
        & =  \mu(i)+\Delta_i-2(2\alpha\beta+\beta)\ci_{[0,1]}\left(Mn_{t}(i^*),\delta\right)\\
        &> \mu(i)+2(2\alpha\beta+\beta)\ci_{[0,1]}\left(Mn_{t}(i),\delta\right)\\
        &\geq  \hat{\mu}_t^j\left(i\right)+(2\alpha\beta+\beta)\ci_{[0,1]}\left(Mn_{t}(i),\delta\right).
    \end{split}
\end{equation*}
It shows the fact that the lower confidence bound of arm $i^*$ is larger than the upper confidence bound of arm $i$,
contradicting the rules of the algorithm to pull arm $i$.

It follows from Equation (\ref{eq:con_2}) that $4(2\alpha\beta+\beta)\ci_{[0,1]}\left(M(n_{t}(i)-1),\delta\right)\geq \Delta_i$, since the algorithm pull arms in a round robin manner.
Last, we apply the confidence interval function for a Bernoulli process, and prove that the number of observations of $i$ received by agent $j$ is upper bounded by
\begin{equation*}
    Mn_t(i)\leq \frac{8(2\alpha+1)^2\beta^2\log \delta^{-1}}{\Delta^2_i}+M.
\end{equation*}



\end{document}